\documentclass[10pt,conference]{IEEEtran}
\IEEEoverridecommandlockouts
\usepackage{cite}
\usepackage{amsmath,amssymb,amsfonts}
\usepackage{graphicx}
\usepackage{textcomp}
\usepackage{xcolor}
\def\BibTeX{{\rm B\kern-.05em{\sc i\kern-.025em b}\kern-.08em
    T\kern-.1667em\lower.7ex\hbox{E}\kern-.125emX}}

\usepackage{balance}
\usepackage{nicematrix}
\usepackage{multirow}
\usepackage{booktabs}
\usepackage{algorithm}
\usepackage{algorithmic}

\usepackage{amsmath}
\usepackage{tcolorbox}
\usepackage{ulem}

\usepackage[numbers]{natbib}

\newcommand{\name}[1]{\textit{Dr. DRL}}

\begin{document}

\title{An Intentional Forgetting-Driven Self-Healing Method For Deep Reinforcement Learning Systems}

\author{
\centering
\IEEEauthorblockN{Ahmed Haj Yahmed$^*$, Rached Bouchoucha$^*$, Houssem Ben Braiek, Foutse Khomh}
\textit{\{ahmed.haj-yahmed,rached.bouchoucha,houssem.ben-braiek,foutse.khomh\}@polymtl.ca}
\IEEEauthorblockA{
\textit{SWAT Lab., Polytechnique Montréal, Canada}}
}

\maketitle
\def\thefootnote{*}\footnotetext{Both authors contributed equally to this research.}\def\thefootnote{\arabic{footnote}}

\begin{abstract}
Deep reinforcement learning (DRL) is increasingly applied in large-scale productions like Netflix and Facebook. As with most data-driven systems, DRL systems can exhibit undesirable behaviors due to environmental drifts, which often occur in constantly-changing production settings. Continual Learning (CL) is the inherent self-healing approach for adapting the DRL agent in response to the environment's conditions shifts. However, successive shifts of considerable magnitude may cause the production environment to drift from its original state. Recent studies have shown that these environmental drifts tend to drive CL into long, or even unsuccessful, healing cycles, which arise from inefficiencies such as catastrophic forgetting, warm-starting failure, and slow convergence. In this paper, we propose \name{}, an effective self-healing approach for DRL systems that integrates a novel mechanism of intentional forgetting into vanilla CL (i.e., standard CL) to overcome its main issues. \name{} deliberately erases the DRL system's minor behaviors to systematically prioritize the adaptation of the key problem-solving skills. Using well-established DRL algorithms, \name{} is compared with vanilla CL on various drifted environments. \name{} is able to reduce, on average, the healing time and fine-tuning episodes by, respectively, 18.74\% and 17.72\%. \name{} successfully helps agents to adapt to 19.63\% of drifted environments left unsolved by vanilla CL while maintaining and even enhancing by up to 45\% the obtained rewards for drifted environments that are resolved by both approaches.
\end{abstract}


\section{Introduction}
\label{sec:introduction}
Deep reinforcement learning (DRL), the blend of deep learning (DL) and reinforcement learning (RL), has shown promising achievements in recent years, such as AlphaStar \cite{arulkumaran2019alphastar}, Netflix’s customized recommendation system \cite{amat2018artwork}, and Facebook ReAgent \cite{gauci2018horizon}.
Nevertheless, DRL-based systems, as with all data-driven systems, are likely to generate incorrect or undesirable behaviors when operating in constantly ever-changing production settings. This phenomenon is known as environmental drift \cite{wang2019incremental}, which is the manifestation of concept drift \cite{padakandla2020reinforcement, webb2016characterizing} in DRL systems. 
The burgeoning field of ML (and RL) operations (MLOps) (and RLOps \cite{li2022rlops}) is establishing the best practices and methods for implementing efficient, trustworthy, and maintainable self-learning systems. Among the most demanding approaches are healing strategies \cite{psaier2011survey, perino2013framework, ghosh2007self} that enable the ML system to recognize its malfunction after a concept shift and to adjust, accordingly, its behaviors in order to regain its performance. Conventional software healing solutions are used to identify and respond to software failures in production by making the appropriate tweaks, at runtime, to restore the system's normal behavior \cite{gazzola2018automatic}. The healing workflow begins with a failure detection step. The healing step follows, which involves executing a healing operation. Finally, a verification step ensures that the software is restored properly. Otherwise, the healing operation is repeated until the failure is resolved or no further action is possible \cite{gazzola2018automatic}. Self-healing solutions \cite{psaier2011survey, gazzola2018automatic} describe techniques that fully automate all steps of healing without any human involvement. During a self-healing operation, a software system should adjust its own structure and behavior in production based on its interpretation of the context, the occurred failure, and its requirements \cite{psaier2011survey}.

Indeed, a DRL system learns its optimal policy (i.e., best actions w.r.t perceived states) by interacting with its environment \cite{sutton1998reinforcement} and by performing reward-driven learning without supervision. Hence, it can be argued that the DRL agent can adapt its policy to handle a newly-shifted environment by engaging in continual learning (CL) cycles for a sufficient amount of time. In practice, various inefficiencies can adversely affect the CL ability of DRL agents, including catastrophic forgetting \cite{french1999catastrophic}, failure of warm-starting \cite{wexler2022analyzing}, Primacy Bias \cite{nikishin2022primacy}, and Capacity Loss \cite{lyle2022understanding}. Thus, as the environment in production drifts from its original state, CL is likely to result in slow healing cycles \cite{mirzadeh2022architecture,khetarpal2022towards} associated with prolonged system dysfunction and downtime. Even worse, a recent study \cite{tonella} reported that CL is confined within environmental parameters beyond which it is unable to heal the DRL agent. Therefore, it is of paramount importance that alternatives and improvements to vanilla CL (i.e., standard CL) be explored, in order to enlarge the adaptation frontiers and accelerate the adaptation curve for low-cost, fast, and effective self-healing DRL systems.

In this paper, we propose \name{}, a self-healing approach for DRL that leverages intentional forgetting \cite{beierle2019intentional, timm2019intentional} combined with continual learning in order to optimize the agent's plasticity and accelerate its adaptation to drifted environments. Intentional forgetting is an interdisciplinary concept that has gained traction in the software engineering community, with applications in cybersecurity \cite{shands2021intentional}, machine learning \cite{beierle2019intentional, reuter2019intentional, timm2018intentional}, and logistics supply networks \cite{timm2019intentional}. Its goal is to deliberately delete unnecessary knowledge in order to focus attention on the system's key components. To apply the intentional forgetting mechanism in DRL self-healing, we adapt the top-k neuron coverage (TKNC) \cite{ma2018deepgauge, xie2019deephunter}, which is a fine-grained layer coverage criterion proposed by DeepGauge \cite{ma2018deepgauge}, in order to identify the major and minor DRL agent behaviors at the level of its neural network units (i.e., neurons). Accordingly, our approach associates the DRL agent's major behavior with hyperactive neurons (i.e., neurons with high levels of activation) and minor behavior with hypoactive neurons (i.e., neurons with low activation levels). Then, our approach intentionally erases the minor behaviors by assigning low-scaled random weights to their associated hypoactive neurons. As a first benefit, forgetting the minor behaviors that are likely to be non-transferable to the drifted environments, increases both the speed and the effectiveness of major behavior adaptation. As a second gain, under-scaling weight initialization as a minor behavior forgetting mechanism enables the self-healing at dual speeds, i.e., major behavior neurons (unchanged) would receive significant and more frequent updates than minor behavior neurons (under-scaled).

To demonstrate the effectiveness of \name{}, we evaluate it on purposefully drifted gym \cite{brockman2016openai} environments with different drifting intensities. Results show that \name{} succeeds in speeding up the self-healing process in terms of runtime and fine-tuning episodes, compared to vanilla CL. \name{} extends the self-healing frontiers that CL provides by 20\% on average. Finally, \name{} is able to maintain and, in some cases, to increase the average reward of CL by up to 45\%.\\
Below, we summarize the contributions of this paper:
\begin{itemize}
   \item[--] We propose the first application of the intentional forgetting concept for the DRL system's self-healing problem. 
    \item[--] \name{} exploits the intricacies of DNN structure to separate the DNN's computation units responsible for major and minor behaviors as seen in the interaction with the original environment.
    \item[--] \name{} leverages the gradient-based optimization principles to smoothly achieve dual-speed self-healing, in which major behaviors are prioritized over minor behaviors.
    \item[--] We conduct the evaluation of \name{} using (i) three popular gym environments that we deliberately drift with varying parameters shifts, namely CartePole \cite{barto1983neuronlike}, MountainCar \cite{moore1990efficient}, and Acrobot \cite{geramifard2015rlpy}; (ii) three well-established DRL algorithms, namely Deep Q-Learning (DQN) \cite{mnih2015human}, Soft Actor-Critic (SAC) \cite{haarnoja2018soft}, and Proximal Policy Optimization (PPO) \cite{schulman2017proximal}. 
    \item[--] In order to stimulate more studies on this topic and promote the undertaking of replication studies, we provide our replication package \cite{contributor_drdrl}.

\end{itemize}
The remainder of this paper is structured as follows. Section \ref{sec:preliminaries} introduces the fundamental concepts that will be used throughout our work. Section \ref{sec:example} provides a concrete illustration of the problem we are addressing and how we are tackling it. Section \ref{sec:approach} describes each step of our approach. Section \ref{sec:evaluation} reports the evaluation outcomes. Section \ref{sec:threats} analyzes the threats to validity. Section \ref{sec:related_work} presents the related works. Finally, Section \ref{sec:conclusion} concludes the paper.

\section{Preliminaries}
\label{sec:preliminaries}
This section briefly introduces DRL, CL, environmental drifts, and self-healing with intentional forgetting.

\subsection{Deep Reinforcement Learning}
Reinforcement Learning (RL)~\cite{sutton1998reinforcement} is a mathematical framework for experience-driven autonomous learning. RL aims to create fully autonomous agents that interact with their environments to learn optimum behaviors and improve over time, through a trial-and-error paradigm ~\cite{sutton1998reinforcement}. Formally, the agent is given a state $s$ at each time-step $t$ of its interaction with the environment, and it decides which action to take based on that state \cite{li2017deep}. Those actions are determined by its on-learning policy, $\pi$, encoded as state-to-action mappings. Once an action is taken, the agent receives a reward and the environment moves to the next state. 

Deep reinforcement learning (DRL), leverages DL to scale RL to complex, high-dimensional state and action spaces \cite{arulkumaran2017deep, li2017deep}. Neural Networks (NN) are non-linear function approximators that are trained to estimate quantities that depend on state and/or actions, e.g., a policy $\pi$ or an action-value function $Q$. DRL algorithms are either model-free or model-based \cite{arulkumaran2017deep}. To avoid modeling biases and solve environments with complex dynamics, we decided to focus our study on model-free DRL algorithms. These algorithms can be divided into three categories \cite{arulkumaran2017deep}. First, \textbf{Policy gradients algorithms} represent The policy with a NN whose weights are updated by maximizing the expected return. This optimization is performed on-policy, meaning that each update is carried out with data coming exclusively from the actual policy. \textbf{Value-based algorithms} represent the action-value function, $Q$, with a NN whose weights are updated based on the Bellman equation \cite{bellman1952theory}. This optimization is performed off-policy, meaning that it leverages data coming from any policy, not only from the actual policy. Finally, \textbf{Hybrid algorithms} use both policy gradients and value-based methods. One notable class of algorithms from this category is the Actor-Critic algorithm which has two networks: Actor and Critic. The actor decides which action should be taken and critics inform the actor how good the action was and how it should adjust. 

\subsection{Environment Drift and Continual Learning}

Standard RL environment settings are often not representative of all possible contexts that agents will encounter for task fulfillment \cite{dulac2019challenges, panzer2022deep}. The behavior of these environments is subject to changes, which can manifest as alterations in real deployment environments or shifts in the parameters of virtual environments \cite{wang2016novel, wang2019incremental}. These challenges, known as partial observability and non-stationarity \cite{dulac2019challenges} in DRL literature, hinder productionizing DRL to real-world problems \cite{dulac2019challenges, panzer2022deep}. For instance, recommendation systems have no observations of the ever-changing mental state of the users and need to always heal their behavior accordingly. These changes in the production RL environment, called environment drift, may affect the performance of trained agents. The environment's behavioral changes become classified as environmental drifts when their severity/intensity is substantial, preventing the trained agent from achieving its task. Although a drifted environment yields different rewards for a subset of state-action pairs, it should not be considered a completely new environment. In fact, the original environment and its drifted counterpart share some regions of state-action space. 
Therefore, the agent in a drift environment only requires healing by fusing new information with existing knowledge (i.e., fine-tuning) rather than learning from scratch. This required fine-tuning is called the continual learning (CL) problem in DL \cite{hadsell2020embracing}, which investigates whether NNs can acquire new knowledge incrementally. In spite of its success in several use cases, it has been shown that new knowledge may interfere with the existing one, which results in replacing it entirely. This failure of stability in CL is called catastrophic forgetting \cite{french1999catastrophic}, in which new experience overwrites previous experience. Thus, the fundamental objective is to design an advanced self-healing approach to improve the adaptability of DRL systems effectively and efficiently.

\section{Motivating Example}
\label{sec:example}

\begin{figure*}[h]
\centering
\includegraphics[width=1\textwidth]{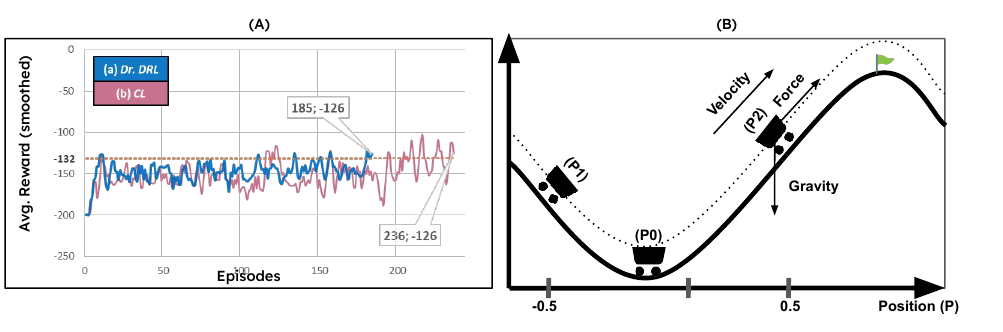}
\caption{(A) The adaptation curves for the DQN agent's average reward per episode on MountainCar: (a) depicts the curve of the average reward when Dr. DRL is used, whereas (b) represents the curve of average reward when CL is used (B) Illustration of the MountainCar environment as described in Moore's Moore's work \cite{moore1990efficient}}
\label{fig:mountaincar}
\end{figure*}

Figure \ref{fig:mountaincar} shows the MountainCar environment \cite{moore1990efficient}, where a car starts on a one-dimensional track situated between two mountains. The objective is to reach the goal position up the right mountain. The car starts at the bottom of the valley, and receives negative rewards, as each timestep passes, to encourage it to reach the goal as soon as possible. 
MountainCar's environment is episodic, where episodes end when $200$ timesteps have passed or when the car reaches the flag. In order to solve the MountainCar, the agent must achieve an average reward of at least $ -110$ over $100$ consecutive evaluation episodes. After drift, we assign 20\% of tolerance in the average reward. So the agent must achieve at least $ -132$.

Drifts in the MountainCar environment can affect three parameters, namely, force, gravity, and goal velocity. Originally, the initial settings of a MountainCar environment, $E$, are set as follows: force$=1e^{-3}$, gravity$=2.5e^{-3}$, and goal\_velocity$=0$. A drift $d$ on $E$, for instance, can shift these parameters to force$=1.2e^{-3}$, gravity$=4e^{-3}$, and goal\_velocity$=0$, leaving the agent unable to solve the drifted environment $E_d$. CL would be used to fine-tune the trained agent on $E_d$ for a set number of episodes. Nevertheless, CL is prone to inefficiencies such as catastrophic forgetting \cite{french1999catastrophic} and slow adaptation, resulting in inferior DRL system performance and higher repair costs. A trained DQN agent, for instance, took $236$ training episodes and $436$ seconds to adapt to $E_d$ leveraging CL (Figure \ref{fig:mountaincar}). Our proposed approach, \name{}, optimizes cost-effectiveness, thereby reducing adaptation expenses. Considering the same $E_d$, \name{} adapted the same DQN agent in $185$ episodes and $273$ seconds. The DQN agent also achieved an average reward of $-122.3$ when adapted with \name{}, compared to $-123.7$ with CL. Finally, as evidenced by Figure \ref{fig:mountaincar}, Dr. DRL improved the stability and monotonicity of the adaptation process by minimizing the fluctuation breadth of the average reward.


\section{Approach}
\label{sec:approach}
In this section, we first describe the problem statement. Then, we introduce our intentional forgetting mechanism for an improved DRL healing. Last, we detail the different phases and steps involved in \name{}'s workflow.
\subsection{Problem formulation}

Let $E$ be a development RL environment, defined as a set of $n$ parameters $\{p_1,. . .p_n\}$, and let $A_E$ be an agent trained on $E$ until it solves $E$ or reaches the desired performance level. Next, let $E'$ represent a new RL environment with parameters $\{p_1',. . .p_n'\}$, denoting the production RL environment.
Although the $A_E$ is trained on $E$, the production RL environment represents the same problem as $E$ and the $A_E$ is expected to be able to solve $E'$ or maintain its performance level as long as the parameters of $E'$ have not drifted substantially from $E$. If an important environmental drift occurs, the performance of $A_E$ decreases proportionally to the deviations, i.e., the greater the drift, the lower the performance. In this case, the RL agent, $A_E$, requires a self-healing mechanism to adapt to the drifted production environment, $E'$, and the healing success depends on its ability to solve $E'$ or regain its optimal level of performance. Conventionally, Continual Learning (CL) is the mainstream approach for self-healing against environmental drifts. Nevertheless, this approach can be inefficient due to several issues the community encounters, such as catastrophic forgetting \cite{french1999catastrophic} and slow adaptation \cite{mirzadeh2022architecture}. In our approach, we assume that a trained agent $A_E$ in the environment $E$ learns both major and minor behaviors, where major behaviors encode critical problem-solving abilities and minor behaviors have no direct impact on problem-solving, but were necessary to achieve the desired performance on the development environment. Structurally, the RL agent encapsulates a neural network, thus, we propose a neuronal categorization method to depict the hyperactive neurons (i.e., contribute almost to all of the predictions) and the hypoactive neurons (i.e., contribute intermittently to the predictions). The proposed neuronal categorization provides a structural representation of the problem-solving skills acquired by the optimal RL agent as a result of interaction with the development environment. Indeed, we believe that the inefficiency of CL may be partially due to the equal importance assigned to both major and minor behaviors, resulting in slow adaptation (all neurons are considered to be adapted) and catastrophic forgetting (CL may reverse the neurons' roles from major to minor, or vice versa). Our proposed self-healing mechanism is an improvement over vanilla CL by (i) prioritizing adjusting major behaviors to maintain the structural representation of problem-solving skills, and (ii) accelerating the behavioral adaptation using dual-speed continuous learning, in which minor behaviors are updated less frequently than major behaviors. These improvements can enhance both the speed and success of self-healing against environmental drifts, which remain unsolved by vanilla CL. In the following, we detail the main components of our approach, along with the self-healing workflow that we have adopted.

\subsection{Intentional Forgetting Mechanism}
The phenomenon of winning lottery tickets has been shown \cite{frankle2018lottery, malach2020proving} to exist in deep neural networks, i.e., there are sub-networks that can be further fine-tuned to achieve as good performance as their source neural network. An example of its implications is the growing trend of reducing the size and resource consumption of DNNs by pruning irrelevant neurons after training~\cite{siswanto2021block}. In line with the lottery ticket hypothesis, the major DRL agent behaviors are likely to be yielded by a sub-network within the agent's DNN. This sub-network encodes the critical problem-solving abilities responsible for achieving the major task. More precisely, this sub-network encapsulates the set of hyperactive neurons (i.e., major-behavior-responsible units) that are expected to represent the essential agent skills to resolve most of the occurring environment states and predict the optimal action. We hypothesize that the DRL agent's major behaviors (i.e., set of hyperactive neural network units) should be primarily adjusted by the DRL self-healing process against a drifted environment, while the minor behaviors are more likely to be non-transferable and should be learned again. Indeed, minor behaviors are likely to have no direct impact on problem-solving but are necessary to achieve the desired performance in a specific environment. Therefore, we adopt the concept of intentional forgetting~\cite{timm2018intentional, beierle2019intentional} that removes irrelevant knowledge to improve the cognitive capabilities of intelligent systems by allowing them to focus solely on pertinent aspects of the given situation. 
In the following, we describe the two main steps of our intentional forgetting mechanism for the enhancement of the DRL agent's healing ability and its associated continual learning stability.\\
\subsubsection{Detection of Minor Behavior Regions}
\label{sec:detct_minor_behavior}
The first step of our intentional forgetting mechanism is to localize the regions of the DRL system responsible for the major behavior (i.e., contributing almost to all of the predictions) and minor behavior (i.e., contributing intermittently to the predictions). Structurally, the RL agent encapsulates a neural network, thus, we propose a neuronal categorization method to depict the hyperactive neurons (i.e., major-behavior-responsible units) and the hypoactive neurons (i.e., minor-behavior-responsible units). This neuronal categorization method provides a structural representation of the problem-solving skills acquired by the RL agent.

A neural network's activations represent the intermediate computations that pass essential information between layers. The activations are computed using a nonlinear function attached to each neuron, which determines whether it should be activated (“fired”) or not, based on the relevance of neuron output for the final prediction.
As illustrated in Equation~\ref{exp_frwd_pass} of the forward pass \cite{bishop1994neural}, all the activations contribute to the last layer activation, $a^{[L]}$.

\begin{small}
\begin{equation}
\label{exp_frwd_pass}
a^{[L]}=\sigma\left(W^{[L]} g\left(\ldots g\left(W^{[1]} a^{[0]}+b^{[1]}\right) \ldots\right)+b^{[L]}\right)
\end{equation}
\end{small}

Where $W = \{W^{[l]}, \forall \,l \in [1 , L ] \}$ and $b = \{b^{[l]}, \forall\, l \in [1, L]\}$ are the weights and biases of a DNN.\\
Hence, the neurons with higher activation levels, called hyperactive neurons, encode the representation of relevant features associated with major behaviors at that processing layer and thus play a significant role in the overall predictions. In contrast, hypoactive neurons, characterized by lower activation levels, encode minor behavior features, but they still contribute to the model's generalizability and confidence. Hyperactive neurons (i.e. neurons with high levels of activation) have been found to provide beneficial learning patterns for DNNs in several empirical studies\cite{ma2018deepgauge, xie2019deephunter}. The concept of identifying hyper-hypoactive neurons was inspired by the Top-K-Neuron-Coverage (TKNC) technique \cite{ma2018deepgauge}. The TKNC was developed to assess the diversity of activation patterns during DNN testing. When computing, TKNC, neurons with higher activation levels are deemed to be more significant in processing the input and are ranked higher, whereas neurons with lower activation are ranked lower. \\ 
Therefore in \name{}, to categorize the hypoactive and hyperactive neurons, we select the $N_H$ hypoactive neurons per layer, i.e., those triggering low activation scores on a regular basis. According to the following formula, $N_H$ is a relative number derived from a prefixed forget rate, $F_r$: $N_{H}^{l} = Round((F_r/100) * N_{l} )$, where $N_H^{l}$ represents the number of hypoactive neurons in layer $l$ and $N_{l}$ is the number of neurons in layer $l$.\\

\subsubsection{Forgetting \& Dual Speed Gradient-based Healing}
\label{sec:dual_speed_healing}
After detecting the major and minor behaviors of the DRL system, \name{} intentionally erases its minor behaviors and proceeds to dual-speed gradient-based healing. We assume that the inefficiency of CL may be partially due to the equal importance assigned to both major and minor behaviors, resulting in slow adaptation (all neurons are equally updated) and catastrophic forgetting (CL may reverse the neurons' roles from major to minor, or vice versa). Hence, our approach offers two improvements to CL. As a first benefit, forgetting the minor behaviors that are likely to be non-transferable to the drifted environments, is expected to increase both the speed and effectiveness of major behavior adaptation. Second, the minor behavior forgetting mechanism enables self-healing at dual speeds, i.e., major behavior units would receive significant and more frequent updates than minor behavior units. Nevertheless, nullifying them would negatively affect the layer's weight asymmetry, and the loss gradients w.r.t null weights would become zero as well; so their corresponding neurons freeze and no longer receive updates. Hence, our proposed method to properly forget the minor behaviors involves the re-initialization of their associated neurons using the original weight initializer. Indeed, random weights are sampled to maintain the asymmetry between neurons and calibrate the values distribution variance proportional to the size of the layer's input. Furthermore, we under-scale these random weights generated for hypoactive neurons using a scale rate, $S_r < 1$. 

\begin{small}
\begin{equation}
\label{exp_back_prop}
\frac{\partial \mathcal{L}}{\partial {z}^{[1]}}=\left[{W}^{[2]}\right]^{\mathrm{T}} \ldots \left[{W}^{[L]}\right]^{\mathrm{T}} \cdot \frac{\partial \mathcal{L}}{\partial z^{[L]}} * q^{[L-1]} * \ldots * q^{[1]}
\end{equation}
\begin{equation}
\label{back_propp}
W^{(i+1)}=W^{(i)}-\eta\: \frac{\partial \mathcal{L}}{\partial W^{[l]}} \quad; \quad b^{(i+1)}=b^{(i)}-\eta\: \frac{\partial \mathcal{L}}{\partial {b}^{[l]}}
\end{equation}
\end{small}

Equations~\ref{exp_back_prop} and ~\ref{back_propp} show how the parameters of the model are updated at the iteration $i+1$, where $\mathcal{L}$ is the loss function; ${z}^{[l]}$ $W^{[l]}$, ${b}^{[l]}$ are respectively the weighted-sum outputs, weights and biases of a layer $l$ and $q^{[l]}= g^{[l] \prime}\left(z^{[l]}\right)$ with $g^{[l]} \prime$  referring to the computed gradients in the layer $l$.

Equation~\ref{exp_back_prop} shows that the gradient-based updates include multiplication by weights, which makes their initial magnitude scale affect their growth and decay over iterations, thus controlling their learning speed. If $S_r$ tends to $0$, new weights would be very small, almost zeros, so their associated neurons would experience low update rates. If $S_r$ tends to $1$, minor behavior neurons will have original scale weights and learn at the same rate as major behavior neurons.\\
Hence, the reduction of weights magnitude affects directly the scale of gradient updates derived by backpropagation through layers. Lowering the magnitude of the reinitialized neurons' weights ensures the slow re-learning of the DRL agent's minor behavior, whereas maintaining the original magnitude of the retained neurons' weights guarantees substantial updates of the RL agent's major behaviors for stable and fast adaptation. Therefore, our proposed intentional forgetting mechanism leverages the weight-gradient scaling relationship to enable dual-speed gradient-based healing of the DRL agent's behaviors.\\


\subsection{Self-healing Workflow}
\begin{figure*}[h]
\centering
\includegraphics[width=1\textwidth]{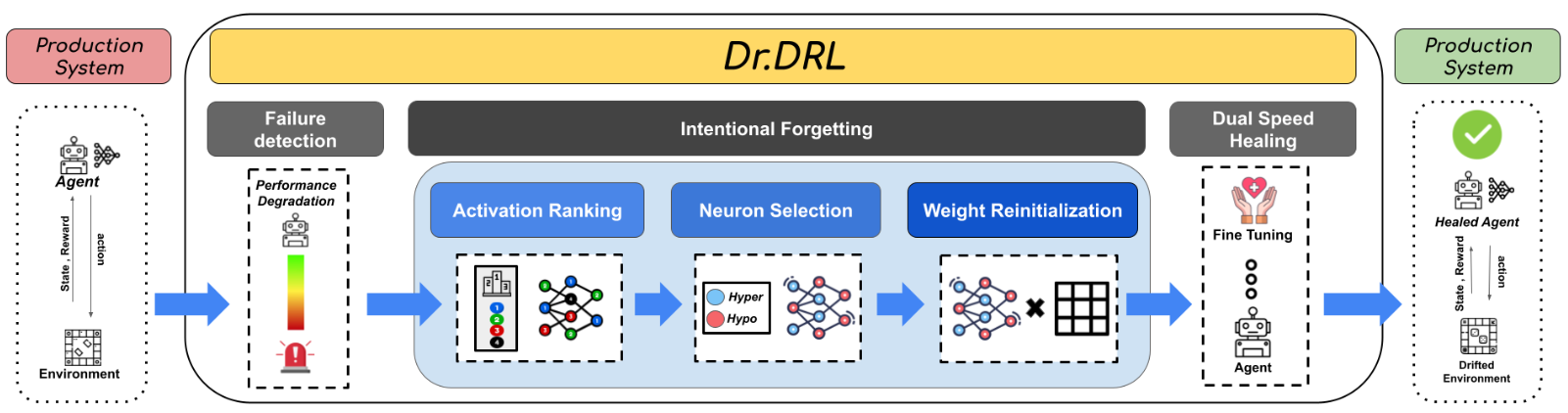}
\caption{Illustration of \name{} Pipeline  
}
\label{fig:pipeline}
\end{figure*}

The overview of our proposed self-healing workflow is illustrated in Figure \ref{fig:pipeline}. Below are details of the workflow's steps.

\textbf{Pre-deployment minor behavior identification}. \name{} requires a pre-deployment step in which it collects the activation traces of the agent’s DNN at the end of development. During the training process, the agent should learn how to map between states and actions properly to maximize the accumulated rewards received in the original environment \cite{sutton2018reinforcement}. Once an optimal DRL agent has been identified, we run it on samples of observations taken from the original training episodes and save the activation traces. 

\textbf{Failure detection}. In \name{}, we develop an event handler component that monitors the production environment and triggers the healing mechanism when an environmental drift occurs. Upon a certain level of parameter shifts, the trained agent loses its ability to predict optimal actions, which means the agent cannot reach the required minimum reward. In that case, the actual agent fails to solve the newly-drifted environment, and consequently, our proposed self-healing mechanism is triggered. 

\textbf{Self-Healing mechanism}. \name{} first identifies the trained agent's minor behavior regions (i.e., hypoactive neurons) using the collected activation traces. In each layer, we rank the neurons based on their activations, and then we select the hypoactive neurons, i.e., triggering lower activations over most of the observations (lines $2$ to $6$). The hypoactive neurons per layer constitute the minor behavior neurons, whose counts are derived according to the prefixed $F_r$ (See Section \ref{sec:detct_minor_behavior}). \name{} then intentionally erases the DRL agent’s minor behaviors and proceeds to dual-speed gradient-based healing. Algorithm \ref{algo:algo1} illustrates the steps of \name{}’s healing mechanism. First, we reinitialize the weights associated with minor behavior neurons using the same weight initializer but under-scaled by $S_r$ in order to prioritize the updates of major over minor behavior neurons (lines $7$ to $11$) (See Section \ref{sec:dual_speed_healing}). Second, we adjust the hyperparameters of the DRL algorithm, especially those that control how much exploration the DRL agent does over the training. We reload the replay memory obtained from previous training episodes on the original environment in order to preserve the initial performance (only for off-policy RL algorithms like DQN and SAC). As soon as the setup is complete, we start fine-tuning the partially-reinitialized DRL agent on the drifted environment (lines $12$ to $14$) (See Section \ref{sec:dual_speed_healing}). The objective is to heal the agent from the involved environmental drifts, i.e., achieve the same level of performance as the original environment.

\begin{algorithm}
 \caption{Dr. DRL's healing mechanism}
 \label{alg:algo1}
 \begin{algorithmic}[1]
 \renewcommand{\algorithmicrequire}{\textbf{Input:}}
 \renewcommand{\algorithmicensure}{\textbf{Output:}}
 \REQUIRE $E'=\{p'_{1}, ..., p'_{n}\}$ drifted environement,  
$A_{E}$: trained agent that solves $E = \{p_{1}, ..., p_{n}\}$, $n_{t}$: training episodes, $obs$: set of training observations, $F_{r}$: Forget rate, $s_{r}$: Scale rate, $init$: weights initializer, $hyp$: hyperparameters of the agent.
 \ENSURE  $A'_{E'}$: agent adapted to the drifted environement $E'$
  \STATE $mask$ :=  $\left [  \right ]$;
  \\ /* \textbf{Detection of Minor Behavior} : /*
  \STATE $acts$ := compute\_activation\_trace($A_{E}$, $obs$)

  \FOR {$i$ in  $A_{E}.hidden\_layers$}
  \STATE hypo\_n := $detect\_minor\_regions( A_{E}, obs, F_{r} )$;
  \STATE masks.append $( generate\_mask(A_{E}), i, hyp, hypo\_n)$;
  \ENDFOR
  \\ /* \textbf{Intentional Forgetting of Minor Behavior} : /*
  \FOR {$i$ in  $A_{E}.hidden\_layers$}
  \STATE w := $apply\_mask( weights\_mask, A_{E})$;
  \STATE new\_w := $reinitialize(w, masks[i], s_{r}, init)$;
  \STATE $A_{E}$.set\_weights($new\_{w}$)
  \ENDFOR
  \\ /* \textbf{Dual Speed Gradient-based Healing} : /*
  \FOR {$eps$ in  $n_{t}$}
  \STATE $A'_{E'}$ := $dual\_speed\_cl(A_{E}, E', hyp)$;
  \ENDFOR
  
 \RETURN $A'_{E'}$ 
 \end{algorithmic} 
 \label{algo:algo1}
 \end{algorithm}

\section{Evaluation}
\label{sec:evaluation}
In this section, we first introduce the DRL case studies (environments and algorithms), as well as our experimental settings, metrics, and procedure. Next, we evaluate \name{} against vanilla CL in terms of cost reduction (RQ1), adaptability enhancement (RQ2), and reward improvement (RQ3).
\subsection{Experimental Setup}
\label{sec:experimental_Setup}
\subsubsection{Environments}
We leverage the following 3 different environments in our empirical evaluation. \\
\textbf{The MountainCar environment} \cite{moore1990efficient} was described in Section \ref{sec:example}. \\
\textbf{The CartPole environment} \cite{barto1983neuronlike} consists of an inverted pendulum attached to a cart that moves on a track controlled by a force. The agent's goal is to keep the pole upright, receiving a reward of $+1$ for each successful time step. The task is episodic, and the episode ends if the pole falls $>15^{\circ}$, the cart moves $>2.4$ units from the center, or 200 time steps pass. To solve the task, an agent must achieve an average reward $>195$ over 100 testing episodes. The CartPole environment is parameterized by four parameters: masspole, length pole, mass cart, and friction.\\
\textbf{The Acrobot environment} \cite{geramifard2015rlpy, sutton1995generalization}  has two joints and two links, and the goal is to raise the lower link to a specified height. If the agent achieves the goal, it gets a reward of $0$, otherwise, it receives a reward of $-1$ at each time step. Episodes end when the goal or $500$ time steps are reached, and the reward threshold for the environment is $-100$ over $100$ evaluation episodes. The environment parameters are `link length 1', `link compos 1', `link mass 1', and `link mass 2'.
\subsubsection{DRL Algorithms}

In the following, we present the selected DRL algorithms for our evaluation. They are widely used and most RL libraries provide a stable implementation of them. \\
\textbf{Deep Q-Learning (DQN)} \cite{mnih2015human, urtans2018survey} is a popular value-based algorithm that leverages neural network to approximate the action-value function, Q. \\
\textbf{Proximal Policy Optimization (PPO)} \cite{schulman2017proximal} is a state-of-the-art policy-based algorithm whose goal is to maximize policy optimization without compromising performance.\\
\textbf{Soft Actor Critic (SAC)} \cite{haarnoja2018soft} is a hybrid DRL algorithm. It focuses on maximizing returns and policy entropy (i.e., degree of its ``randomness"), simultaneously.\\ 

\subsubsection{Software and Hardware Configuration}
We implemented our approach as an open-source tool using Python 3.7~\cite{van2009python} and it supports Tensorflow (version 2.4.4)~\citep{abadi_tensorflow}. We evaluated environments from the Gym library (version 0.23.1)~\cite{brockman2016openai} and DRL algorithms adapted from popular GitHub repositories \cite{noauthor_welcome_nodate, TF2RL}. We used two GPU-enabled servers. The first server runs Ubuntu 20.04 and features a 24-core AMD EPYC 7413 CPU, an ASPEED AST2500 GPU, and 500 GB of RAM. The second runs CentOS 7 and has a 12-core 1.70 GHz Intel Xeon Bronze CPU, an NVIDIA GeForce RTX 2080 Ti GPU.
\subsubsection{Evaluation Procedure and Metrics}\hfill\\
\noindent{\textbf{Procedure:}} The DRL algorithms' hyperparameters were tuned using the open-source repository rl-baselines-zoo~\citep{stable_baselines}. As minor implementation changes may substantially affect a DRL algorithm performance~\citep{engstrom2020implementation}, we conducted ten distinct training runs with various random seeds. The final hyperparameters for each DRL algorithm were chosen based on the highest average reward. For each DRL algorithm, we sampled six realistic environmental drift settings. To simulate different shifts in production, we intentionally induce drifts of varying magnitudes. For distinguishing between intensities of simulated drifts, we rely on the environment's parameter ranges described in~\citep{tonella}. Then, we use either our approach, \name{}, or vanilla CL to heal the DRL agent to the six newly-drifted environments. To have a fair comparison, we used the same hyperparameters and max number of episodes of the DRL agent beforehand. Nevertheless, we tuned the specific parameters, $F_r$ and $S_r$, of \name{} for each environment using, respectively, the ranges of [50, 40, 30, 20, 10]\% and [0.00001, 0.0001, 0.001, 0.01, 0.1]. We found that the best parameters are $F_r$=50\% and $S_r$=0.1 for the CartPole environment. For both Acrobot and MountainCar environments, $F_r$=10\% and $S_r$=0.0001 work better. Each time, $F_r$ and $S_r$ values were kept the same across all neural network layers. We hypothesize that the depth and type of layer may have an effect on the values of $F_r$ and $S_r$. We did not thoroughly examine this topic in our study, and it could serve as the focus of future studies. 

The difference in the parameters $F_r$ and $S_r$ of \name{} across CartPole, Acrobot and MountainCar can be explained by the level of difficulty. The easiest environment here is CartPole. MountainCar and Acrobot pose more challenges due to their deceptive natures. In fact, these deceptive reward environments discourage agents from exploring (by offering a negative reward), causing them to fall into local optimality. Following the guidance of the prior study~\citep{tonella}, we assign an adaptability tolerance ratio of 20\% to these two environments. As a result, we accept a performance degradation of 20\% during the adaptation. For instance, in the MountainCar environment, the agent adapts to the drifted environment if it gets an average reward of $-132$, corresponding to $-110$ (solved environments) minus 20\% (see Figure~\ref{fig:mountaincar}(A)).\\
Finally, all experiments are run 10 times to overcome the stochastic nature of these DRL algorithms. We use statistical significance testing, i.e., the non-parametric Wilcoxon~\citep{wilcoxon_individual_1945} test and the Vargha-Delaney A12 effect size~\citep{vargha2000critique}, to compare the results obtained by \name{} and vanilla CL.\\
\textbf{Environmental Drifts:} On the basis of the study~\cite{tonella} on the plasticity of DRL agents, two types of drifted environments are identified: 1) an adaptable environment to which the agent will be able to adapt, and 2) a non-adaptable environment to which the agent cannot adapt. Hence, we consider both types of drifted environments. For the adaptable type, our goal is to evaluate the stability of our self-healing approach, as well as its ability to heal the behavior of the RL agent faster and more economically. Regarding the non-adaptable type, we aim to investigate how well our self-healing approach improves the DRL agent's adaptability (healing ability) and enables its adaptation despite the failure of CL.\\
\textbf{Metrics:} Below, we introduce the different evaluation metrics that have been used in the empirical evaluations.\\ 
\textit{Increase Ratio (\%IR(X)) / Decrease Ratio (\%DR(X)).} It consists of the percentage increase/decrease of the quantity X from applying \name{} rather than CL, as formulated in Eq.~\ref{IR}/Eq.~\ref{DR}. Thereby, positive values indicate the on-watch quantity, X, has known an improvement, whereas negative values indicate a degradation.\\
\begin{equation}
\label{IR}
\%IR(X) = \dfrac{\textit{\name{}\_X} - \textit{CL\_X}}{\textit{CL\_X}} \times 100
\end{equation}

\begin{equation}
\label{DR}
\%DR(X) = \dfrac{\textit{CL\_X} - \textit{\name{}\_X}}{\textit{CL\_X}} \times 100
\end{equation}

\textit{Adaptability Ratio (\%AR).} It consists of the percentage of pairs (a trained DRL agent, a drifted environment) that a given approach Y is able to adapt, as formulated in Eq.~\ref{AR}.

\begin{equation}
\label{AR}
\%AR(Y) = \dfrac{ \textit{(Env, Agent)}_{adpt_{(Y)}}}{ \textit{(Env, Agent)}_{All}} \times 100
\end{equation}

\subsection{Research Questions and Answers}

\subsubsection{RQ1 (Cost-effectiveness): How cost-effective \name{} is in adapting DRL systems compared to continual learning?} \hfill  

\textbf{Motivation:} The purpose of this RQ is to compare \name{} with the vanilla CL technique in terms of the cost-effectiveness of healing the DRL systems that failed to solve drifted environments. As the healing strategies operate on the DRL agent at runtime, delays in healing cycles can be resource-intensive and costly, and consequently, such operations are constrained by a maximum number of steps allowed in solving episodic environments. Thus, the more effective a healing strategy is, the shorter the time it takes.\\
\noindent{\textbf{Method:}} We measure the time and episode count necessary to heal a DRL system in response to a drifted environment. Both metrics are important to compare the cost-effectiveness of \name{} and CL because one healing strategy can solve the drifted environment in fewer episodes while it reaches the maximum number of steps allowed in each episode. Hence, the time reflects the total number of steps done by the healing strategy. To consider a healed DRL agent as having successfully solved a drifted environment, it must achieve an average reward over 100 consecutive episodes greater or equal to the defined threshold (i.e., same as the original environment). To estimate the magnitude of the speed-up obtained when using \name{} versus vanilla CL, we compute the decrease ratio, \%DR (Eq.~\ref{DR}), of the fine-tuning time and episodes reached by \name{} over the ones obtained by vanilla CL. 

\begin{table}[]
    \caption{Comparison of Healing Time and Fine-Tuning Episodes Decreases Using \name{} over vanilla CL.}
    \label{tab:time}
    \centering
        \begin{NiceTabular}{ll|rrr}
    \CodeBefore
    \Body
        \toprule
        & \textbf{} &   \textbf{DQN} &     \textbf{PPO} &     \textbf{SAC} \\
        \midrule
        
       \multirow{3}{*}{\textbf{Episodes DR (\%)}} 
        & \textbf{CartePole}  &     $\boldsymbol{21.2}$  & \underline{$\boldsymbol{45.9}$}  & 0.9  \\
        & \textbf{MountainCar}  &     17.7  & \underline{$\boldsymbol{2.9}$}  & $\boldsymbol{20.1}$  \\
        & \textbf{Acrobot}   &     3.2  & 9.3  & \underline{$\boldsymbol{38.3}$}  \\
        \midrule
        \multirow{3}{*}{\textbf{Time DR (\%)}} 
        & \textbf{CartePole}  &     \underline{$\boldsymbol{30.3}$}  & $\boldsymbol{-20.5}$  & 1  \\
        & \textbf{MountainCar}  &     $\boldsymbol{20.3}$  & \underline{$\boldsymbol{24.4}$}  & \underline{$\boldsymbol{18.7}$}  \\
        & \textbf{Acrobot}   &     5.3  & 16.2  & \underline{$\boldsymbol{73}$}  \\
        
        \bottomrule
\end{NiceTabular}
\end{table}

\begin{figure}[h]
\centering
\includegraphics[width=0.5\textwidth]{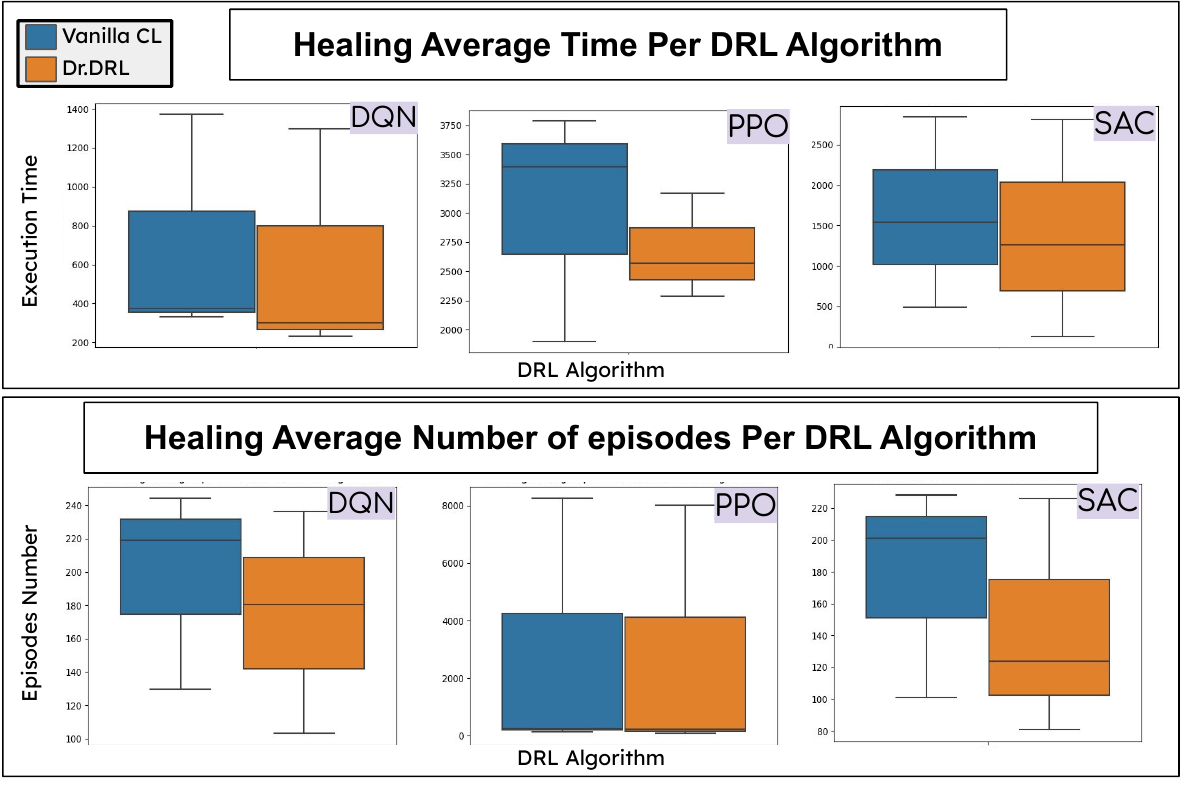}
\caption{Comparison of Healing Time and Fine-Tuning Episodes Box Plots Using \name{} over vanilla CL.}
\label{fig:time_plot}
\end{figure}

\noindent{\textbf{Results:}} Table \ref{tab:time} shows that \name{} often requires fewer healing cycles as measured by both fine-tuning time and episode count than its conventional alternative, vanilla CL. Indeed, the bold values indicate that the difference between the two compared healing approaches is statistically significant (i.e., the p-value $< 0.05$). The underlined values indicate that their associated magnitude of the effect size A12 is medium or large. Furthermore, Figure \ref{fig:time_plot} demonstrates the juxtaposition of box plots drawn from the distributions of time and episode count obtained by \name{} versus vanilla CL when tried with different pairs of DRL algorithms and environments. In line with statistical tests, the median values of both time and episode count achieved by \name{} are consistently lower than their counterparts yielded by vanilla CL.

To be more specific, \name{} outperforms vanilla CL in terms of fine-tuning episode count across all the pairs of agents and environments, where the gap between them were statistically significant in $44\%$ of experiments with either a medium or large effect size. In regards to healing time, \name{} outperforms vanilla CL across all the pairs of agents and environments, except for the pair of "PPO" algorithm and "CartePole" environment. In fact, \name{} succeeds in healing PPO-based agents against drifted CartePole environments using longer steps by episode than its conventional alternative, vanilla CL, which delayed the healing time by \name{} even with very few fine-tuning episodes.


\begin{tcolorbox}[colback=blue!5,colframe=blue!40!black]
\textbf{Finding 1.} \name{} enables a cost-efficient healing of DRL systems in drifted environments, through fewer fine-tuning episodes and shorter healing time. 
\end{tcolorbox}

\subsubsection{RQ2 (adaptability frontiers): Can \name{} enhance the adaptability of DRL systems to drifted environments?} \hfill

\textbf{Motivation:} Biagiola and Tonella \cite{tonella} demonstrated that vanilla CL has an environmental drift boundary beyond which it fails to heal the DRL system. \name{} is designed with the objective of expanding the constrained scope of adaptable drifted environments. Hence, we assess the ratio of drifted environments that cannot be resolved through CL, but are instead resolved by \name{}, and vice versa.

\noindent{\textbf{Method:}} We compute the adaptability ratios (\%AR) (Eq.~\ref{AR}) obtained by the two compared healing approaches, \name{} and vanilla CL, for all the studied agent-environment pairs. We expect these ratios can be divided into four major groups, representing four possible scenarios: (1) agent-environment pairs fitted by both healing approaches (i.e., \name{} and vanilla CL); (2) pairs fitted exclusively by \name{}; (3) pairs fitted exclusively by vanilla CL; and (4) pairs not fitted by both approaches.

\begin{figure*}[h]
\centering
\includegraphics[width=0.7\textwidth]{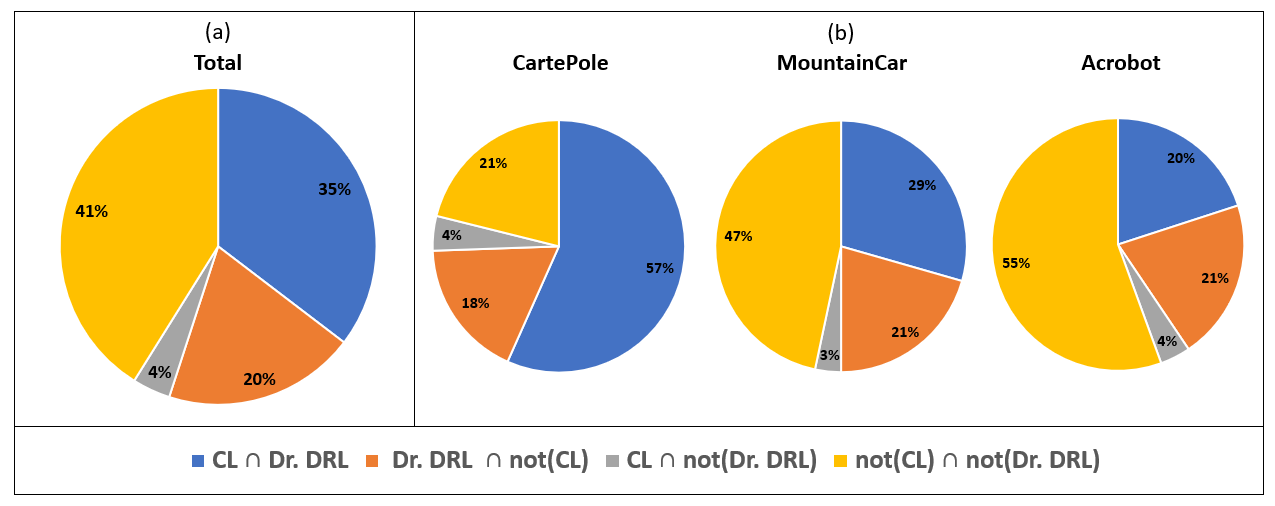}
\caption{Distribution of \%AR over the 4 identified groups Per healing Method and RL Environment}
\label{fig:pie_chart}
\end{figure*}

\noindent{\textbf{Results:}} Figure \ref{fig:pie_chart} shows (a) the Pie chart demonstrating the global adaptability ratios (\%AR) achieved by either \name{}, vanilla CL, both, or none of them; and (b) Pie charts depicting the same results of adaptability ratios (\%AR) broken down by environments.

As can be seen in Figure \ref{fig:pie_chart}-a, our approach, \name{}, succeeds in extending the boundary of adaptable environment ratio obtained by vanilla CL. More specifically, \name{} was able to heal about 20\% of agent-environment pairs on which vanilla CL fails. This shows the effectiveness of intentional forgetting of minor behaviors in the enhancement of the DRL system self-healing ability, leading to a reduced number of non-adaptable environmental drifts. Nevertheless, this expansion in the adaptability frontiers of vanilla CL using \name{} was achieved at the cost of a few drifted environments on which \name{} fails despite they can be resolved by vanilla CL. The reason is that any intentional forgetting, no matter how carefully screened or guided, induces a loss of information (i.e., the reset hypoactive neurons). The latter may cause inefficiencies in the DRL self-healing ability. However, the results of RQ1 shows the cost-effectiveness resulting from our intentional forgetting mechanism in terms of the sample complexity. Hence, these 4\% of non-adaptable drifted environments can be down to zero by reducing the ratio of erased minor behavior neurons over the healing trials by \name{} in its future versions (i.e., the less minor behavior neurons are reset, the more \name{} behaves as a vanilla CL). 

In line with the previous study on DRL system plasticity \cite{tonella}, there are environmental drift frontiers of the DRL system adaptation in response to drifted environments that are beyond the capabilities of both vanilla CL and our novel healing approach, \name{}. This reinforces the assertion that the continuous self-healing of in-production DRL agents to constantly-shifting environments has limits. There are still instances of drifted environmental that require the re-training of the DRL agent from scratch or the redesign of the DRL algorithm beforehand in order to be able to resolve them.

According to Figure \ref{fig:pie_chart}-b, the proportions of agent-environment pairs that are healed exclusively by one of the two healing approaches remain almost constant across all the studied RL environments. Meanwhile, the proportions of pairs that are either adapted or not equally by both approaches differ substantially depending on the studied environment. This can be explained by the differences in the parameterization complexity of these environments, i.e., involving high dimensional and sensitive parameters. Drifted instances derived from complex parametric environments are more challenging for self-healing approaches. 

Research in healing software systems is very active and continues to advance towards equipping modern systems with self-healing capabilities in order to cope with ever-changing requirements throughout operation \cite{perino2013framework}. For DRL systems, our proposed approach, \name{}, is in line with these current efforts, and by increasing the healing success ratio from 39\% (i.e., 35\% + 4\%) to 55\% (i.e., 35\% + 20\%), it represents a step further in assuring the DRL system's self-healing ability in response to environmental drifts often encountered in production settings.\\


\begin{tcolorbox}[colback=blue!5,colframe=blue!40!black]
\textbf{Finding 2.} \name{} outperforms the vanilla CL in healing a higher proportion of DRL systems to drifted environments, expanding the vanilla CL adaptability frontiers.
\end{tcolorbox}

\subsubsection{RQ3 (Performance): Are \name{}'s healed systems more performant in terms of average reward than their vanilla CL counterparts?} \hfill

\noindent{\textbf{Motivation:}} Predictive performance remains the main selection criterion for any DRL system. Thus, restoring or even exceeding the original performance in drifted environment after healing is an important factor to evaluate a DRL system's self-healing approach. Hence, we conduct a comparison between \name{} and vanilla CL in terms of the resulting healed DRL system's predictive performance in drifted environments.


\noindent{\textbf{Method:}} We compute the average of rewards obtained by the healed agent over the course of 100 consecutive episodes, using the two compared self-healing approaches, \name{} and vanilla CL. Then, we calculate \name{}'s average reward increase ratio, \%IR (Eq.~\ref{IR}), w.r.t the vanilla CL's average reward. Using the above-mentioned metrics, we solely consider the two following scenarios: (i) jointly adaptable environments, i.e., the agent-environment pairs on which vanilla CL and \name{} both succeeded; (ii) jointly non-adaptable environments, i.e., the agent-environment pairs on which CL and \name{} both failed. The remaining scenarios, where one approach, either vanilla CL or \name{}, succeeds and the other fails, are irrelevant for predictive performance comparison because the gap between them will be wide due to the failure of one of them in healing the DRL system.

\begin{table}[]
    \caption{Comparison of healed DRL Agent Reward Increases Using Dr.DRL over CL.}
    \label{tab:reward}
    \centering
        \begin{NiceTabular}{ll|rrr}
    \CodeBefore
    \Body
        \toprule
        & \textbf{} &   \textbf{DQN} &     \textbf{PPO} &     \textbf{SAC} \\
        \midrule
        
       \multirow{3}{*}{\textbf{Reward IR (\%)}} 
        & \textbf{CartePole}  &     -0.1  & -0.06  & \underline{$\boldsymbol{-0.2}$}  \\
        & \textbf{MountainCar}  &     0.8  & -0.5 & $\boldsymbol{10.4}$  \\
        & \textbf{Acrobot}   &     -0.5  & -1.2  & \underline{13.1}  \\
        \midrule
        \multirow{3}{*}{\textbf{Reward IR (\%)}} 
        & \textbf{CartePole}  &     \underline{$\boldsymbol{-34.9}$}  & \underline{45.7}  & 2.2  \\
        & \textbf{MountainCar}  &     \underline{3.5}  & \underline{$\boldsymbol{25.8}$}  & -15.4  \\
        & \textbf{Acrobot}   &     \underline{$\boldsymbol{14.2}$}  & 13.5  & $\boldsymbol{21.2}$  \\
        
        \bottomrule
\end{NiceTabular}
\end{table}

\begin{figure*}[h]
\centering
\includegraphics[width=0.8\textwidth]{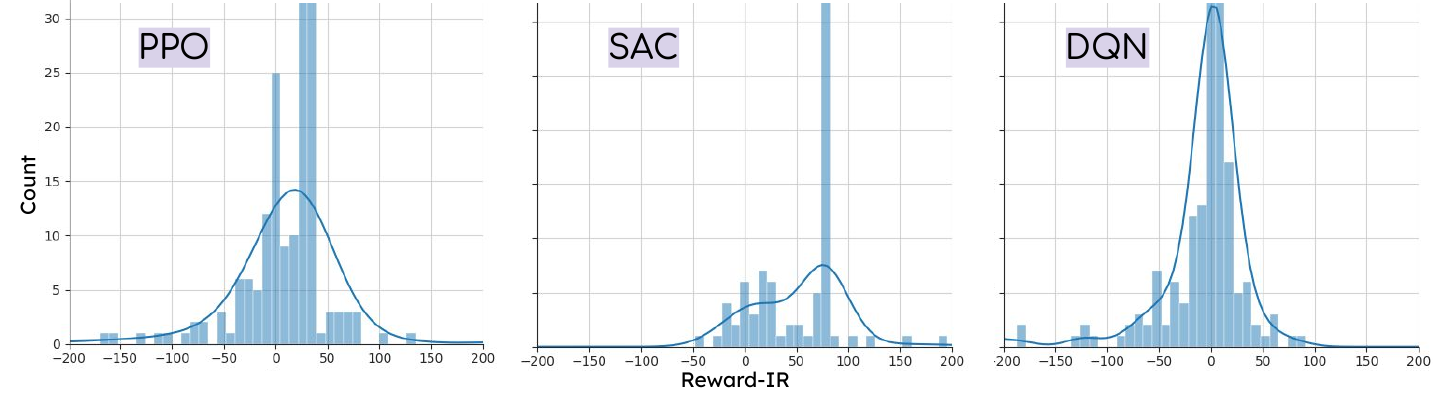}
\caption{Comparison of non-adaptable environments average Reward Increase Using Dr.DRL over CL.}
\label{fig:reward_plot}
\end{figure*}

\noindent{\textbf{Results:}} The first row of Table \ref{tab:reward} shows the reward increase ratio of \name{} compared to vanilla CL in jointly adaptable environments. \name{} increases the average reward obtained in 22\% of experiments by at least 10\%, while maintaining an almost equal average reward (no less than -1.2\%) in the remaining 78\% of the experiments. The second row of Table \ref{tab:reward} shows the reward increase ratio of \name{} compared to vanilla CL in jointly non-adaptable environments. \name{} was able to improve the average reward in 78\% of the configurations by up to 45\% of the increase ratio over vanilla CL's average reward. Table \ref{tab:reward} demonstrates that \name{} first succeeds in improving or at least maintaining the baseline average reward achieved by CL in jointly adaptable environments. Second, \name{} meets average rewards better than vanilla CL in drifted environments which both self-healing approaches cannot resolve.
For a more detailed view, we show the Figure \ref{fig:reward_plot}, containing the distributions of \name{}'s reward increase ratio (averaged by DRL algorithms) over vanilla CL in jointly non-adaptable environments. Overall, distributions follow a curve that tends to skew towards the right (increase ratio $\geq 0$). These left-skewed distributions (also called right-leaning curves) show that our DRL system's self-healing approach, \name{}, is able to exceed vanilla CL's predictive performance in non-adaptable environments.

\begin{tcolorbox}[colback=blue!5,colframe=blue!40!black]
\textbf{Finding 3:} \name{} was capable of achieving similar or even higher performances in adaptable environments, and it also earned higher performances in non-adaptable ones, showing how much closer \name{} was to meet the requirements than vanilla CL.
\end{tcolorbox}

\section{Threats to Validity}
\label{sec:threats}
In the following, we discuss the threats to validity.\\
\textbf{Threats to internal validity.} They may result from how the empirical study was conducted. To mitigate these issues, we have configured both \name{} and CL with identical parameters (e.g., number of training episodes, maximum allowed steps per episode, etc.) across all studied pairs of RL environments and DRL algorithms.\\
\textbf{Threats to conclusion validity.} They are often related to random variations and inappropriate use of statistical tests. To mitigate these threats, we ran each experiment using either \name{} and CL, $6\times10$ times ($6$ environments with varying parameter shifts, and $10$ different optimized DRL agents). Specifically, the training of 10 agents of each DRL algorithm using different random seeds, was intended to overcome the DRL algorithm's randomness. To confirm the statistical significance of our findings, we performed statistical hypothesis testing and effect size assessments using the non-parametric Wilcoxon test~\cite{wilcoxon_individual_1945} and the Vargha–Delaney effect size~\cite{vargha2000critique}.\\
\textbf{Threats to external validity.} They may limit the generalizability of our findings to other RL-based systems. Although more subjects would be needed to fully assess the generalizability of our results, we have chosen 3 classic and diverse environments that are widely used in the DRL community, as well as, their implementations are included in gym library \cite{brockman2016openai}. In addition, we have assessed our method on three distinct DRL algorithms from different types and having distinct learning strategies.\\
\textbf{Threats to reproducibility.} In order to guarantee the reproducibility of our evaluation, we offer a comprehensive replication package \cite{contributor_drdrl}. This package encompasses the complete source code of Dr.DRL, along with all necessary information and hyperparameters essential for replicating our assessment. This includes explicit details regarding the architecture of the tested agents and their corresponding neural networks. Moreover, in section \ref{sec:experimental_Setup} provide a thorough explanation of the fine-tuning procedures undertaken and the precise values of critical hyperparameters, notably $S_r$ and $F_r$, essential for accurately reproducing our evaluation process.

\section{Related work}
\label{sec:related_work}
This section summarizes related studies to our work.\\
Software healing approaches \cite{gazzola2018automatic} have been explored to fix software system issues ``in the field", as opposed to software repair solutions \cite{monperrus2018automatic}, which are used ``in-house" for bug-fixing operations. Indeed, healing approaches respond to software failures in production by making the appropriate adjustments, at runtime, to restore the system's normal behavior \cite{gazzola2018automatic}. When these adjustments are fully automated without human intervention, the approaches are called software self-healing \cite{psaier2011survey, perino2013framework, ghosh2007self}.

Due to their data-driven nature and unique architecture, specific healing, and repairing approaches have been proposed for ML systems. Rusu et al.~\cite{rusu2016progressive} suggest freezing changes to previously trained networks and introducing additional sub-networks to learn new jobs. Shin et al.~\cite{shin2017continual} train a generative model to create data with the same original data distribution. Thus, original data may be mixed with new data to preserve current knowledge and develop new skills. Next, the learning without forgetting (LwF)~\cite{li2017learning} approach limits network predictions on past tasks from changing while optimizing new task parameters. The above approaches focused on fine-tuning the model using more data. Yet, the fine-tuned model may not remove undesired behavior or efficiently learn new behaviors without catastrophic forgetting~\cite{french1999catastrophic, rolnick2019experience}. Other approaches tried to address the problem without model retraining. Apricot~\cite{zhang2019apricot} is one attempt to fix DNN iteratively through a weight-adaptation method. Additionally, Arachne~\cite{sohn2019search}, a search-based DNN repair method, modified weights directly instead of retraining. Finally, Stocco and Tonella \cite{stocco2022confidence} proposed  a novel CL method for misbehavior DNN predictors in self-driving cars. When data distribution shifts, this technique updates predictors using in-field confidence metric selection and error-based weighted retraining. 

Despite all these attempts, DRL-specific healing still unstudied. Instead, few SE research such as Biagiola and Tonella \cite{tonella} examined the confidence boundary of CL. Biagiola and Tonella \cite{tonella} proposed AlphaTest, an approach for characterizing the DRL's adaptability in its environment. AlphaTest takes a DRL agent that has been trained in a parameterized environment as input and samples the environment parameter space. It then trains the agent in CL mode on the sampled environment configurations, with the goal of characterizing the agent's adaptability frontier. Leveraging AlphaTest,  developers can learn about the healing boundaries of CL.

Unlike prior research, We propose a DRL-specific self-healing approach that uses intentional forgetting \cite{shands2021intentional} and CL to optimize the agent's adaptability. The proposed \name{} could benefit the DRL community since no advanced self-healing approaches have yet been proposed. Instead, traditional healing alternatives are used like training a new policy from scratch or using vanilla CL. 

\section{Conclusion}
\label{sec:conclusion}
In this paper, we propose, \name{}, a novel DRL self-healing approach, which involves an intentional forgetting step to systematically prioritize the adaptation of a DRL agent's problem-solving skills when an environmental drift occurs in production settings. Indeed, \name{} identifies the agent's minor behaviors by watching the neurons with low-ranked activations over training episodes, called hypoactive neurons. If a drift in the environment happens and the agent becomes unable to solve it, \name{} heals the DRL agent using two steps: (i) re-initializes the weights associated with hypoactive neurons with low-scaled values; (ii) performs continual learning updates of the patched DRL agent on the newly-drifted environment. By doing that, \name{} enables the adaptation of DRL agent's behaviors at dual speeds, i.e., untouched major behavior neurons maintain larger updates compared to under-scaled minor behavior neurons. In order to evaluate our proposed improvement, we conduct self-healing experiments of different DRL agents on drifted environments with varying parameter shifts using \name{} and vanilla continual learning. The results demonstrate that \name{} achieves (i) faster healing of the DRL agent, (ii) a higher number of adaptable environments, and (iii) equal or more elevated average rewards than vanilla CL. As part of our future work, we intend to explore other neuron categorization strategies besides activation-based ranking.
\balance

\section*{Acknowledgment}
\label{sec:ack}
This work is funded by the Fonds de Recherche du Quebec (FRQ), the Canadian Institute for Advanced Research (CIFAR), and the National Science and Engineering Research Council of Canada (NSERC). However, the findings and opinions expressed in this paper are those of the authors and do not necessarily represent or reflect those organizations/companies.

\bibliographystyle{IEEEtran}
{\footnotesize
\bibliography{refs.bib}}
\end{document}